\documentclass{article}
\usepackage{MAGE}

\usepackage{times}
\usepackage{soul}
\usepackage{url}
\usepackage[hidelinks]{hyperref}
\usepackage[utf8]{inputenc}
\usepackage[small]{caption}
\usepackage{graphicx}
\usepackage{amsmath}
\usepackage{amsthm}
\usepackage{booktabs}
\usepackage{algorithm}
\usepackage{algorithmic}
\usepackage{colortbl} 
\usepackage[switch]{lineno}
\usepackage{stfloats}
\usepackage{amssymb}
\usepackage{authblk}

\title{MAGE: A Multi-stage Avatar Generator with Sparse Observations}

\author[1]{Fangyu Du}
\author[1]{Yang Yang}
\author[2]{Xuehao Gao}
\author[1]{Hongye Hou}
\affil[1]{School of Automation Science and Engineering,Xi'an Jiaotong University}
\affil[2]{School of Automation,Northwestern Polytechnical University}
\affil[ ]{\{zjdfy2019,houhongye2001\}@stu.xjtu.edu.cn, yyang@mail.xjtu.edu.cn,gaoxuehao.xjtu@gmail.com}

\begin{document}
\maketitle

\begin{abstract}

Inferring full-body poses from Head Mounted Devices, which capture only 3-joint observations from the head and wrists, is a challenging task with wide AR/VR applications. Previous attempts focus on learning one-stage motion mapping and thus suffer from an over-large inference space for unobserved body joint motions. This often leads to unsatisfactory lower-body predictions and poor temporal consistency, resulting in unrealistic or incoherent motion sequences. To address this, we propose a powerful \textbf{M}ulti-stage \textbf{A}vatar \textbf{GE}nerator named \textbf{MAGE} that factorizes this one-stage direct motion mapping learning with a progressive prediction strategy. Specifically, given initial 3-joint motions, MAGE gradually inferring multi-scale body part poses at different abstract granularity levels, starting from a 6-part body representation and gradually refining to 22 joints. With decreasing abstract levels step by step, MAGE introduces more motion context priors from former prediction stages and thus improves realistic motion completion with richer constraint conditions and less ambiguity. Extensive experiments on large-scale datasets verify that MAGE significantly outperforms state-of-the-art methods with better accuracy and continuity.
\end{abstract}

\section{Introduction}

\begin{figure}
    \centering
    \includegraphics[width=\linewidth]{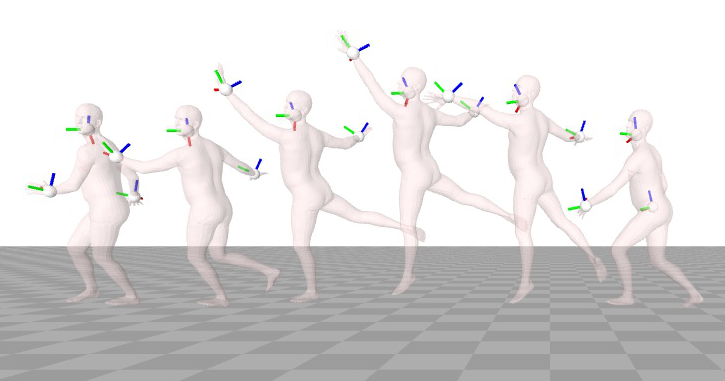}
    \caption{Generating full-body motion from HMDs' observations. The RGB axes represent the motion information of the head and both wrists, serving as the input to our model for generating full-body motion sequences.}
    \label{fig:enter-label}
\end{figure}
With the rapid proliferation of AR/VR technologies and the emergence of various consumer products, there is a growing demand for generating avatars from sparse observations captured by these devices. Conventional systems \cite{jiangTransformerInertialPoser2022,yangLoBSTrRealtimeLowerbody2021a} typically monitor the position, velocity, and orientation changes of Head Mounted Displays (HMDs) and hand controllers to animate the user’s upper-body movements. While these methods can accurately reconstruct upper-body motion, they fail to provide a complete full-body representation, which is crucial for enhancing user immersion and is indispensable in scenarios such as motion training or third-person gaming. One straightforward approach to achieving full-body tracking is to add multiple Inertial Measurement Unit (IMU) sensors like what \cite{huangDeepInertialPoser2018,jiangTransformerInertialPoser2022} do, but this can lead to increased discomfort, higher costs, and more complex calibration procedures. Therefore, it is of great significance to develop methods that can reconstruct the user’s entire body motion from such sparse observations, balancing accuracy with user comfort.\par

In the task of generating full-body motion sequences from sparse observations, both regression-based~\cite{jiangAvatarPoserArticulatedFullBody2022a,zhengRealisticFullBodyTracking2023} and generative approaches~\cite{castilloBoDiffusionDiffusingSparse2023,duAvatarsGrowLegs2023} have shown promising performance. Recently, diffusion models~\cite{hoDenoisingDiffusionProbabilistic2020a,nicholImprovedDenoisingDiffusion2021,sohl-dicksteinDeepUnsupervisedLearning2015} have facilitated improved results, particularly in conditional settings. In this task, the SMPL~\cite{loperSMPLSkinnedMultiperson2015} model is commonly used to describe full-body motion. The motion of each joint in SMPL is determined by its relative rotation to its parent node in the kinematic tree, meaning that joint information is propagated hierarchically through the tree structure. Consequently, predicting motion for joints far from the input nodes suffers from cumulative multi-level errors with a large inference space. Hence, it's challenging to generate lower-body motion due to upper-body inputs. Furthermore, constrained by the single-step 3-to-22 mapping, existing methods struggle to maintain both accuracy and temporal consistency in the generated motion, which is quite important for the quality of the generated results.\par

To address the aforementioned challenges, we propose a multi-scale representation of human motion. Specifically, we iteratively merge adjacent joints in the SMPL model into coarser components. When applied to the generation process, this idea is utilized in reverse: starting with the coarsest representation to establish the overall motion structure, and then progressively adding finer details. Through coarse-grained motion representations, we capture holistic motion information that provides constraints and guidance for subsequent stages. This design offers multiple opportunities for error correction, mitigating cumulative errors by ensuring the accuracy of coarser body parts, particularly at distal joints far from the inputs. Moreover, fewer nodes in coarser stages simplifies relationship modeling, allowing the model to focus on primary motion dynamics and minimize noise propagation. As a result, our method produces more accurate and continuous human motion. \par
Building on this hierarchical idea, we further propose a multi-stage neural network based on a diffusion model, dubbed MAGE. MAGE partitions the task of full-body motion generation from sparse inputs into three sequential phases: the coarsest level first establishes the overall motion trajectory, the second level focuses on refining section-specific movements, and the final phase incorporates all SMPL joints to achieve fully-detailed motion sequences. Throughout these stages, the coarser representations not only serve as a guidepost to constrain subsequent refinements but also ensure global consistency as local details are gradually introduced.\par
To validate our approach, we conduct experiments on the large-scale AMASS~\cite{mahmoodAMASSArchiveMotion2019} benchmark. MAGE outperforms state-of-the-art methods across multiple metrics, demonstrating its efficiency. It not only achieves higher accuracy but also greatly reduces motion jitter while preserving natural motion patterns. Specifically, MAGE improves Mean Per Joint Rotation Error (MPJRE) by 5\%, Mean Per Joint Velocity Error (MPJVE) by 10\%, and Jitter by 11\%.\par
We summarize our contributions as follows:
\begin{itemize}
    \item We propose a multi-scale human motion generation framework that captures motion information at different granularities. Motion is generated progressively from coarse to fine, capturing both global and local information. This hierarchical framework reduces SMPL's intrinsic cumulative errors, especially at distal joints.
    \item We introduce MAGE, a multi-stage generative diffusion model that implement the above strategy. MAGE generates motion from sparse observations in three stages, consisting of 6, 11, and 22 nodes, respectively. Coarser results guide and constrain the later training process by transmitting temporal information and reducing the inference space, making MAGE highly effective.
    \item Our experimental results on large mocap benchmark demonstrate that MAGE achieves state-of-the-art performance in various scenarios for sparse-input human motion generation, effectively balancing the trade-off between accuracy and coherence.
\end{itemize}

\section{Related Work}
\subsection{Motion Tracking from Sparse Inputs}
Recent advancements~\cite{vonmarcardSparseInertialPoser2017,huangDeepInertialPoser2018,yiTransPoseRealtime3D2021} in human full-body motion tracking from sparse inputs have attracted significant attention from researchers, yielding effective and innovative outcomes. Specifically, the Sparse Input Processor (SIP)~\cite{vonmarcardSparseInertialPoser2017} utilized heuristic methods to address this challenge, while the Deep Input Processor (DIP)~\cite{huangDeepInertialPoser2018} was the pioneer in integrating neural networks, employing a bi-directional LSTM to accurately predict the joints of the SMPL manikin. Following these developments, the Physical Input Processor (PIP)~\cite{yiPhysicalInertialPoser2022} and the Tensor Input Processor (TIP)~\cite{jiangTransformerInertialPoser2022} enhanced performance by incorporating physical constraints and selecting alternative base models. These methods have proven the feasibility of deriving full-body motion from sparse IMU inputs. Likewise, LobSTr~\cite{yangLoBSTrRealtimeLowerbody2021a} successfully captured full-body motion using just four IMU inputs—head, dual wrists, and pelvis. However, the widespread adoption of AR/VR technologies poses new challenges, as most devices track only three positions: head and two wrists. To address this limitation, recent studies have proposed new techniques for full-body motion tracking using three inputs. Among these, AvatarPoser~\cite{jiangAvatarPoserArticulatedFullBody2022a} employed a transformer-based architecture, and AvatarJLM~\cite{zhengRealisticFullBodyTracking2023} introduced a joint-level feature to enhance joint interaction modeling, achieving improved results. Additionally, generative approaches like VAEHMD~\cite{dittadiFullBodyMotionSingle2021}, which utilizes a Variational AutoEncoder (VAE)~\cite{kingmaAutoEncodingVariationalBayes}, and FLAG~\cite{aliakbarianFLAGFlowBased3D2022}, which employs normalizing flows~\cite{rezendeVariationalInferenceNormalizing2015}, have been explored. Recent studies leveraging the Diffusion model's superior conditional generation capabilities~\cite{castilloBoDiffusionDiffusingSparse2023,duAvatarsGrowLegs2023,fengStratifiedAvatarGeneration2024}, have also shown promising results.\par
The aforementioned methods have significantly advanced the field of capturing human motion from sparse inputs and reconstructing full-body motion. However, these methods often require more IMU inputs than typically available in practical scenarios or inadequately generate the full-body motion sequence with low accuracy and smoothness.
\subsection{Diffusion Models and Human Motion Generation}

Diffusion models~\cite{sohl-dicksteinDeepUnsupervisedLearning2015,hoDenoisingDiffusionProbabilistic2020a,nicholImprovedDenoisingDiffusion2021} have recently emerged as powerful generative frameworks that progressively add noise to data and then learn to invert this noising process, producing high-fidelity samples. Initially demonstrating their effectiveness in image generation tasks, these models often exhibit greater training stability and superior performance compared to traditional GANs~\cite{dhariwalDiffusionModelsBeat2021}. With deeper research, it proves that diffusion models have outstanding performance especially on conditional generation tasks.\par
In the realm of human motion synthesis, earlier work relied heavily on sequence-to-sequence networks~\cite{fragkiadakiRecurrentNetworkModels2015} and graph-based architectures~\cite{jainStructuralRNNDeepLearning2016} to predict future motions. Although these approaches showed promising results, GAN-based methods emerged to further enhance the realism of generated motions. However, these methods typically need inputs from all body joints---an assumption that proves challenging in many real-world scenarios. More recently, research has shifted towards conditional motion generation, driven by various kinds of conditions, such as textual prompts~\cite{nicholGLIDEPhotorealisticImage2021,guoTM2TStochasticTokenized2022,gaoGUESSGradUallyEnriching2024a}, audio cues~\cite{liAIChoreographerMusic2021,liDanceFormerMusicConditioned2022,aristidouRhythmDancerMusicDriven2023}, or explicit controller constraints~\cite{starkeLocalMotionPhases2020}, achieving significant breakthroughs. Yet, such rich conditioning signals are often unavailable in typical AR/VR applications, where head-mounted devices (HMDs) often provide only 3-joint sparse observations. This limited sensor input necessitates more specialized solutions.\par
Hence, considering diffusion models' outstanding performance in conditional generation tasks, researchers have tried to utilize diffusion models to generate full-body motion from sparse observations, which have demonstrated advanced results. Nevertheless, most existing approaches attempt a direct 3-to-22 joint mapping in a single stage~\cite{duAvatarsGrowLegs2023} or in a single scale~\cite{fengStratifiedAvatarGeneration2024}, often leading to unsatisfactory results and overfitting. To address this limitation, we propose a multi-stage diffusion framework that can utilize different scales' motion information, where earlier stages establish global motion patterns, thereby guide and constrain subsequent refinement stages. This progressive approach effectively alleviates the under-constrained nature of sparse-input motion generation, yielding more accurate and coherent results.

\section{Method}

This section outlines our proposed MAGE network. According to our introduced multi-scale human motion framework, we employ a multi-stage diffusion model to gradually generate human motion sequence, aiming to achieve more reliable and effective outcomes.
\subsection{Problem Formulation}
Our goal is to reconstruct the full-body motion sequence using the sparse observations. After processing and enhancing the observations, they are input into our model to gain a 22-joint motion features which can guide the skinning procedure of the SMPL model so that we can get the generated avatar.  \\
\textbf{Input information.} In this paper, we use observed joint features \(\mathbf{C}^{1:N}\) as the input to the network, where N denotes time steps. In time step n, we gain rotation 
\(\mathbf{R}^n_{1\colon M}\), angular velocity \(\boldsymbol{\Omega}^n_{1\colon M}\), position \( \mathbf{p}^n_{1\colon M}\), and linear velocity \( \mathbf{v}^n_{1\colon M}\) from the original observations like what \cite{jiangAvatarPoserArticulatedFullBody2022a} do, where M denotes the number of observed joints, 
\( \mathbf{R}^n_m \) is represented as a 3×3 rotation matrix, and \( \mathbf{p}^n_m \) is directly obtained as a 1×3 vector.\par
The angular velocity can be calculated as:
\begin{equation}
    \boldsymbol{\Omega}^n=[\mathbf{R}^{n-1}]^{-1}\mathbf{R}^n,
\end{equation}
and the linear velocity can be calculated as:
\begin{equation}
    \mathbf{v}^n=\mathbf{p^n}-\mathbf{p^{n-1}}.
\end{equation}\par
Considering 6D representation's better continuity, we represent rotation and angular velocity in 6D~\cite{zhouContinuityRotationRepresentations2019}, denoted as \({\mathbf{r}^n_{1\colon M}}\) and 
\(\boldsymbol{\omega}^n_{1\colon M}\), respectively. Therefore, we get \(
\mathbf{C}^n=\{\mathbf{r}^n_{1\colon M}, \boldsymbol{\omega}^n_{1\colon M}, \mathbf{p}_{1\colon M}^n, \mathbf{v}_{1\colon M}^n\}\in\mathbb{R}^{(6+6+3+3)\times M}=\mathbb{R}^{18\times M}
\), and \(\mathbf{C}^{1:N}\in\mathbb{R}^{N\times18\times M}\)\\
\textbf{The Outputs.} For human pose description, we employ the SMPL model~\cite{loperSMPLSkinnedMultiperson2015}, focusing on the pelvis and the relative rotation of each joint. We follow~\cite{dittadiFullBodyMotionSingle2021} to exclude facial and hand joints in the skeleton of SMPL model, resulting in the final prediction model covering the first 22 joints only. During inference, we use the model's local rotational predictions to generate body movements, then adjust for the head's translation to determine global movement, integrating these results to model comprehensive human body motion~\cite{jiangAvatarPoserArticulatedFullBody2022a}. Thus, the target of 3D body avatar generation task comes down to predict the first 22 joints of SMPL model which can be denoted by \(\mathbf{X}_0^{1:N}\in\mathbb{R}^{N\times6×\times22}=\mathbb{R}^{N\times132}\).
\subsection{Multi-scale Human Motion Framework}
\begin{figure}[b]
    \centering
    \includegraphics[scale=0.36]{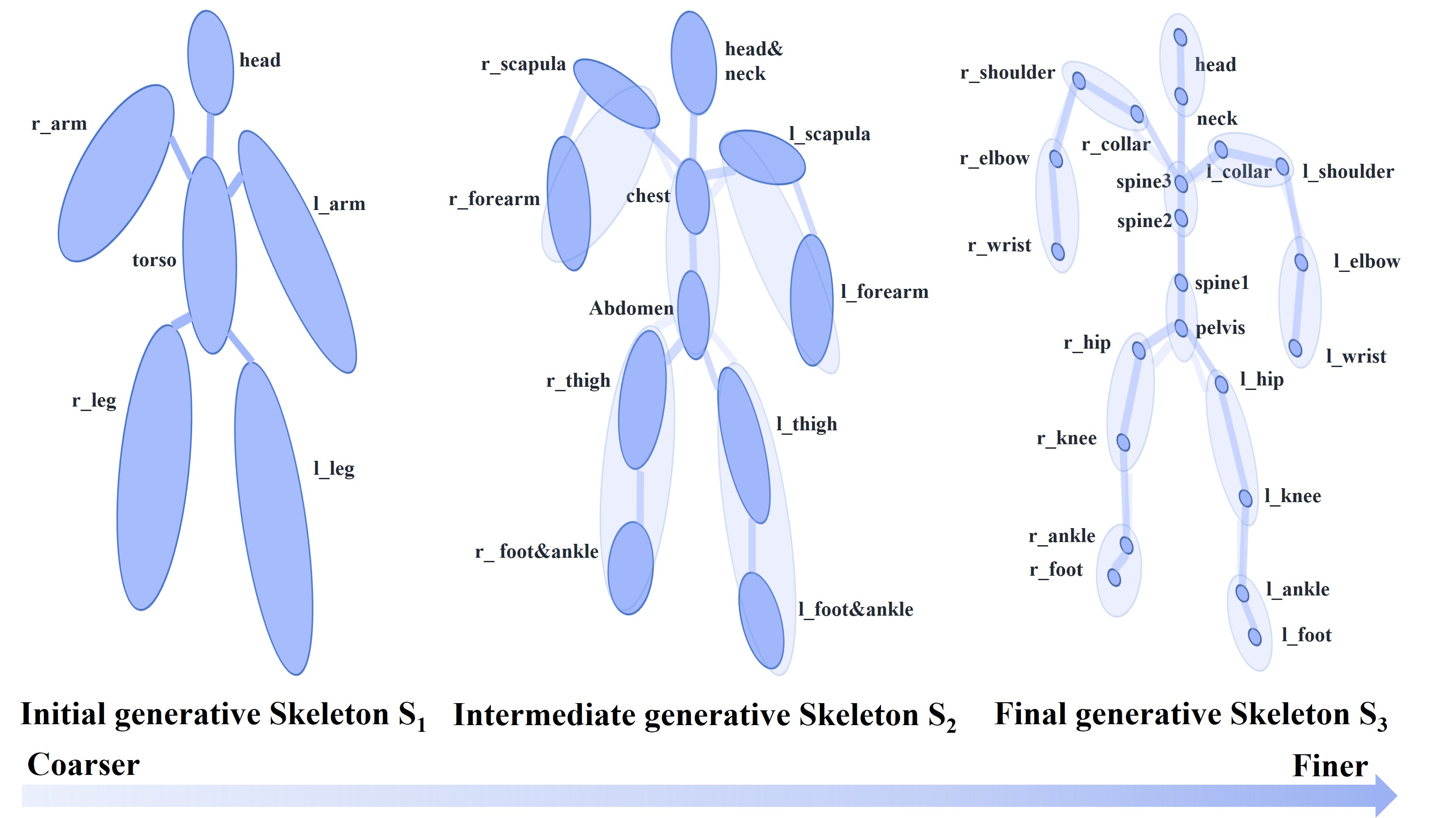}
    \caption{Three body scales from coarse to fine. $\mathbf{S}_1$, $\mathbf{S}_2$, $\mathbf{S}_3$ contain 6, 11 composite nodes and 22 joint nodes, respectively. }
    \label{multiscale}
\end{figure}
In the task of generating full-body motion sequences from sparse observations of the head and the two wrists, most existing methods generate the finest motion sequence directly. However, due to the inherent parent-child node connection in the SMPL kinematic tree, nodes farther away from the input 3 nodes suffer from more error accumulation during the generation process. At the same time, the direct generation from 3 to 22 nodes introduces an overly large inference space, making the method more prone to overfitting, thereby reducing its generalization performance. To address these issues, we propose a multi-scale human motion representation and use it to gradually reconstruct avatar motion. As shown in Figure~\ref{multiscale}, we adopt a three-scale representation: (1) Human Skeleton $\mathbf{S}_1$ with 6 composite nodes as the coarsest representation, (2) Human Skeleton $\mathbf{S}_2$ with 11 composite nodes as an intermediate state, and (3) Human Skeleton $\mathbf{S}_3$ with 22 joint nodes as the final, finest-grained representation of the human skeleton motion, which are the same as what SMPL model use to construct the 3D body avatar.\par
Employing this multi-scale framework, we first generate a coarse-grained motion and then refine it. The coarser motion can not only capture the approximate orientation and position of the entire body but also pay more attention to the temporal consistency. And fewer nodes reduce the propagation of individual errors, helping capture more accurate temporal information. Conditioned on the overall motion information, our framework can reconstruct detailed local motion more accurately and more smoothly in later refinement stages. Each intermediate representation provides new constraints and guidance for producing the next, finer level of detail, which can produce better results and narrow the inference space to reduce the risk of overfitting. Experimental results confirm the feasibility and efficiency of this strategy. Meanwhile, our method further improves model interpretability, making the learning process more intuitive and systematic.
\begin{figure*}[t]
    \centering
    \includegraphics[scale=0.63]{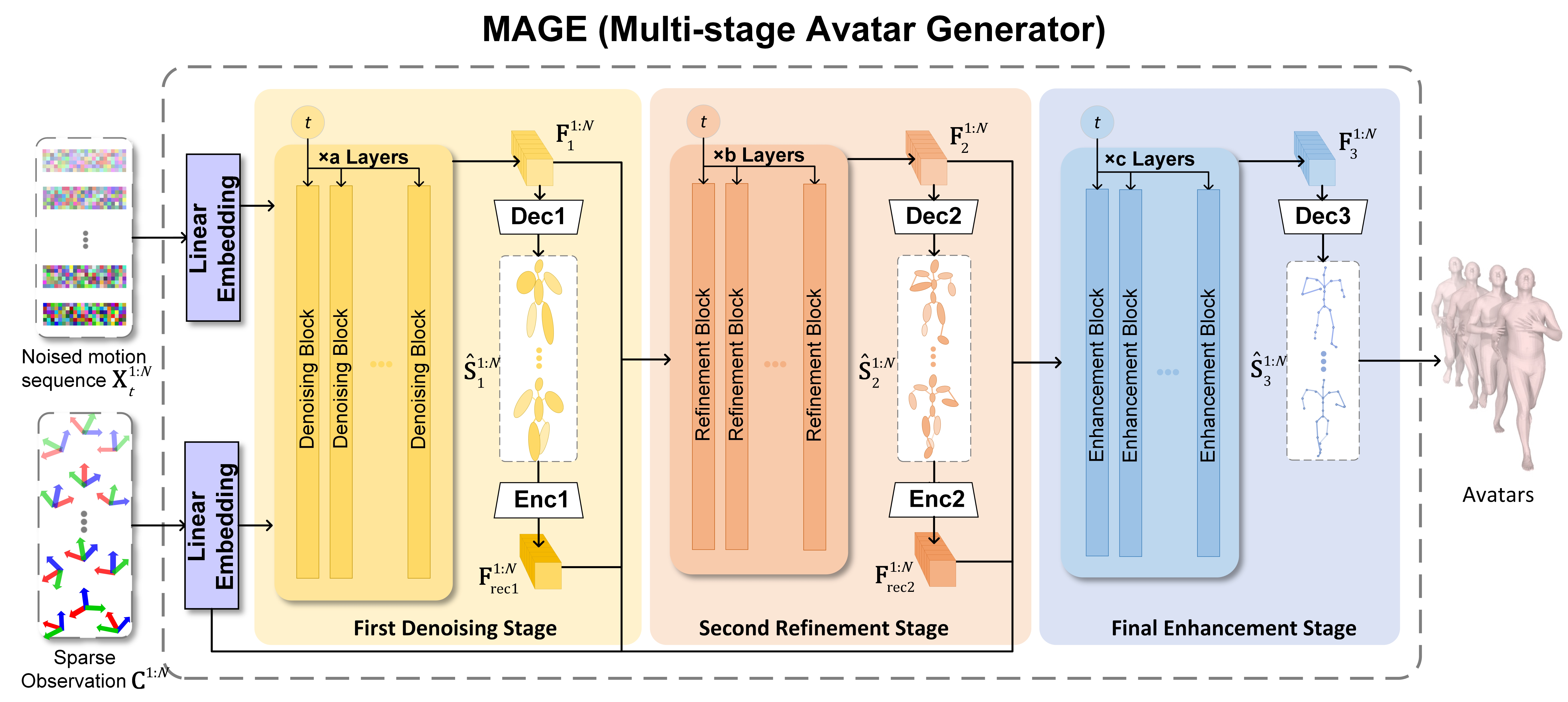}
    \caption{The overall structure of MAGE. We utilize a sparse observation sequence and a full-body motion sequence with t steps of noise addition as inputs to the model. MAGE sequentially generates multiscale full-body motion sequences $\hat{\mathbf{S}}^{1\colon N}_1$, $\hat{\mathbf{S}}^{1\colon N}_2$, and $\hat{\mathbf{S}}^{1\colon N}_3$, where earlier stages' outputs can guide and constrain the training of subsequent stages.}
    \label{pipeline}
\end{figure*}

\subsection{Multi-stage Diffusion Model}
The diffusion model has emerged as a highly popular and efficient generative model in recent years. It operates by simulating the diffusion process in non-equilibrium thermodynamics, gradually transforming random Gaussian noise into the desired data. This is achieved through a learning process that iteratively adds noise and then denoises. In this task, the target data corresponds to the local rotations of the 22 joints in the SMPL model which can be denoted as $\mathbf{X}^{1\colon N}_0$.\par
The forward diffusion process refers to the progression of time steps from $t=0$ to 
$t=T$, during which noise is incrementally added to the original data $\mathbf{X}^{1\colon N}_0$ according to a variance schedule $\beta_1, . . . , \beta_T$. At time steps $T$, we view it as random Gaussian noise $\mathbf{X}^{1\colon N}_T$. This process can be represented by the following probability distribution function:
\begin{equation}
    q(\mathbf{X}_t^{1:N} \mid \mathbf{X}_{t-1}^{1:N}) := \mathcal{N}\big(\mathbf{X}_t^{1:N} ; \sqrt{1-\beta_t} \mathbf{X}_{t-1}^{1:N}, \beta_t I\big).
\end{equation}\par
Conversely, the reverse diffusion process occurs from $t=T$ to $t=0$. In this phase, the model learns to progressively remove noise from the random Gaussian noise $\mathbf{X}^{1\colon N}_T$ to reconstruct the target data $\mathbf{X}^{1\colon N}_0$. This process is described by the following probability distribution function:
\begin{equation}
    p_\theta(\mathbf{X}_{t-1}^{1:N} \mid \mathbf{X}_{t}^{1:N}) := \mathcal{N}\big(\mathbf{X}_{t-1}^{1:N} ; \boldsymbol{\mu}_\theta(\mathbf{X}_{t}^{1:N},t),\beta_t I\big),
\end{equation}
where the mean $\boldsymbol{\mu}_\theta(\mathbf{X}_{t}^{1:N},t)$ can be reformulated~\cite{hoDenoisingDiffusionProbabilistic2020a} as:
\begin{equation}
    \boldsymbol{\mu}_\theta(\mathbf{X}_{t}^{1:N},t) = \frac{1}{\sqrt{\alpha_t}} \big( \mathbf{X}_t^{1:N} - \frac{\beta_t}{\sqrt{1 - \bar{\alpha}_t}}\boldsymbol{\epsilon}_\theta(\mathbf{X}_{t}^{1:N},t)\big),
\end{equation}
where $\alpha_t=1-\beta_t$, $\bar{\alpha}_t=\prod_{i=1}^{t}{\alpha_i}$.\par
In essence, the goal of the diffusion model is to learn how to predict the noise $\boldsymbol{\epsilon}_\theta(\mathbf{X}_{t}^{1:N},t)$ from $\mathbf{X}_{t}^{1:N}$ at any given time step t and computes the denoised output $\mathbf{X}_{t-1}^{1:N}$. Through iterative application of this process, the model finally reconstructs the target data $\mathbf{X}^{1\colon N}_0$ from the initial Gaussian noise $\mathbf{X}_{T}^{1:N}$.\par
For our work specifically, our diffusion model is a conditional generative model that leverages observed joint features \(\mathbf{C}^{1:N}\) as conditions to guide the model, making the reverse diffusion process of the model described as $p_\theta(\mathbf{X}_{t-1}^{1:N} \mid \mathbf{X}_{t}^{1:N},\mathbf{C}^{1:N})$. Unlike traditional approaches that predict the noise $\boldsymbol{\epsilon}_\theta(\mathbf{X}_{t}^{1:N},t)$, we follow~\cite{rameshHierarchicalTextConditionalImage2022} to directly predict the target data $\mathbf{X}^{1\colon N}_0$ from any $t$, which yields a better denoising effect. We adopt the multi-scale human motion framework to divide the denoising process into three stages. In the first denoising stage, MAGE reconstruct $\hat{\mathbf{S}}^{1\colon N}_1$ to capture holistic motion. In the second refinement stage, it generates $\hat{\mathbf{S}}^{1\colon N}_2$ to add more detail. And finally, it outputs $\hat{\mathbf{S}}^{1\colon N}_3$ with 22 joints, which represents the ultimate $\mathbf{X}^{1\colon N}_0$ that we aim to recover.
As shown in Figure~\ref{pipeline}, we embed the noised motion sequence $\mathbf{X}^{1\colon N}_t$ at time step $t$ and the observed joint features $\mathbf{C}^{1:N}$, then concatenate them as the model input. After passing through the denoising modules which are comprised of MLP layers enhanced with RepIn~\cite{duAvatarsGrowLegs2023} as the time-step embedding, the model produces the preliminary denoised latent features  
$\mathbf{F}_1$. These features are then passed through a fully connected layer to generate the 6-component human motion sequence $\hat{\mathbf{S}}^{1\colon N}_1$, which is further embedded into a higher-dimensional representation $\mathbf{F}_{rec1}$ through another fully connected layer. By concatenating $\mathbf{C}^{1:N}$,$\mathbf{F}_1$, and $\mathbf{F}_{rec1}$, the output is fed into the second stage. Subsequent stages follow a similar process to the first stage.
Ultimately, the model defines three objective functions corresponding to $\mathbf{S}^{1\colon N}_1$, $\mathbf{S}^{1\colon N}_2$, and $\mathbf{S}^{1\colon N}_3$ as follows:
\begin{equation}
    L_1 = \mathbb{E}_{\mathbf{X}^{1\colon N}_0\sim q(\mathbf{X}^{1\colon N}_0), t} \left[ \left\| \mathbf{S}^{1\colon N}_1 - \hat{\mathbf{S}}^{1\colon N}_1 \right\|_2^2 \right],
\end{equation}
\begin{equation}
    L_2 = \mathbb{E}_{\mathbf{X}^{1\colon N}_0 \sim q(\mathbf{X}^{1\colon N}_0), \hat{\mathbf{S}}^{1\colon N}_1, t} \left[ \left\| \mathbf{S}^{1\colon N}_2 - \hat{\mathbf{S}}^{1\colon N}_2 \right\|_2^2 \right],
\end{equation}
\begin{equation}
    L_3 = \mathbb{E}_{\mathbf{X}^{1\colon N}_0 \sim q(\mathbf{X}^{1\colon N}_0), \hat{\mathbf{S}}^{1\colon N}_2. t} \left[ \left\| \mathbf{S}^{1\colon N}_3 - \hat{\mathbf{S}}^{1\colon N}_3 \right\|_2^2 \right].
\end{equation}\par
A weighted sum of these objective functions is computed to form the final objective function:
\begin{equation}
    L_{obj} = \alpha L_1+\beta L_2 +\gamma L_3,
\end{equation}
where $\alpha,\beta,\gamma$ are hyperparameters to control the weights of three stages' losses.
\section{Experiments}
\subsection{Implementation Details}
We use rotation (6D), angular velocity (6D), position (3D) and linear velocity (3D) of head, left wrist and right wrist in global coordinate system to consist condition \(\mathbf{C}^{1:N}\in\mathbb{R}^{N\times18\times3}\) to guide the reverse diffusion process. And we use local rotation of first 22 joints in SMPL model to be the target of MAGE, which can be denoted as \(\mathbf{X}_0^{1:N}\in\mathbb{R}^{N\times6\times22}\). Considering the balance between accuracy and continuity, we set $N=120$ in this paper.\par
We set the latent dimension to be 512 and all shapes of latent features in MAGE are $120\times512$. We use 12 denoiser blocks in each stage ($a=12, b=12, c=12$) to guarantee the sufficiency of training. We directly feed the intermediate results $\mathbf{F}_1$ and $\mathbf{F}_2$ into fully connected layers to obtain $\hat{\mathbf{S}}^{1\colon N}_1$ and $\hat{\mathbf{S}}^{1\colon N}_2$ with shapes $120\times36$ and $120\times66$, respectively, instead of first predicting features with a shape of $120\times132$ and then obtaining $\hat{\mathbf{S}}^{1\colon N}_1$ and $\hat{\mathbf{S}}^{1\colon N}_2$ through the previously mentioned mapping from 132 dimensions to 36 and 66 dimensions. We set max time steps $T=1000$ in training and utilize DDIM~\cite{songDENOISINGDIFFUSIONIMPLICIT2021} technique to sample only 4 steps rather than 1000 steps to save plenty of time during inference. Moreover, we use a straightforward yet effective overlapping generation strategy for producing a 120-frame full-body motion sequence, where we add 12 historical frames in to ensure the coherence of the generated motion, as well as to maintain an appropriate speed.\par
On a single NVIDIA V100 GPU, our proposed MAGE model achieves real-time performance by generating a single-frame output in just 0.36 ms using 4-step DDIM sampling, corresponding to an impressive 2778 FPS. This far exceeds the frame rate requirements for AR/VR applications, demonstrating its capability for real-time generation.

\subsection{Dataset and Evaluation Metrics}
\textbf{Dataset.} We conduct both model training and inference on the AMASS dataset~\cite{mahmoodAMASSArchiveMotion2019}, which is a large-scale collection combining multiple motion capture datasets based on SMPL model~\cite{loperSMPLSkinnedMultiperson2015}. For fair comparison, we follow the previous works to use two subsets of AMASS, referred to as $\text{D}_1$ and $\text{D}_2$. 
$\text{D}_1$ follows the scheme proposed by~\cite{jiangAvatarPoserArticulatedFullBody2022a}, randomly splitting CMU~\cite{AMASS_CMU}, BMLr~\cite{AMASS_BMLrub}, and HDM05~\cite{AMASS_HDM05} into training and test sets with a ratio of 90\% for the training set and 10\% for the test set. Meanwhile, $\text{D}_2$ is based on some recent research and adopts a larger subset composed of CMU~\cite{AMASS_CMU}, MPI Limits~\cite{AMASS_PosePrior}, Total Capture~\cite{AMASS_TotalCapture}, Eyes Japan~\cite{AMASS_EyesJapanDataset}, KIT~\cite{AMASS_KIT-CNRS-EKUT-WEIZMANN}, BioMotionLab~\cite{AMASS_BMLrub}, BMLMovi~\cite{AMASS_BMLmovi}, EKUT~\cite{AMASS_KIT-CNRS-EKUT-WEIZMANN}, ACCAD~\cite{AMASS_ACCAD}, MPI Mosh~\cite{AMASS_MoSh}, SFU~\cite{AMASS_SFU}, and HDM05~\cite{AMASS_HDM05} for training, while HumanEval~\cite{AMASS_HumanEva} and Transition~\cite{mahmoodAMASSArchiveMotion2019} serve as the test set. This design aims to evaluate the generalization capability of the model under varying data distributions.\\
\textbf{Evaluation Metrics.} We evaluate the quality of model-generated results from two aspects: static accuracy and dynamic continuity. The former determines whether the generated avatar's position and posture are correct, reflecting the model's ability to predict the 3D human motion state at a single time step. This is a common and important evaluation criterion in 3D human motion generation tasks. The latter determines whether the generated motion is stable and smooth, reflecting the model's consistency in predicting the entire sequence. In 3D human motion generation, particularly in VR and AR applications, continuity largely determines the user experience, which we prioritize.\par
Therefore, for prediction accuracy, we use mean per joint rotation error (MPJRE) and mean per joint position error (MPJPE) as evaluation metrics, and for continuity, we adopt mean per joint velocity error (MPJVE) and Jitter, which indicates the average jerk (the time derivative of acceleration). Additionally, we track the average position error of the root joints (Root PE), hand joints (Hand PE), upper-body joints (Upper PE), and lower-body joints (Lower PE) to pinpoint the strengths and weaknesses of the model in predicting different body regions.
\subsection{Quantitative and Visualized Results}
In dataset $\text{D}_1$, we compare the performance of MAGE with several state-of-the-art methods across the eight metrics presented in Table~\ref{tab:pro1}. Notably, MAGE achieves the best performance on all metrics, indicating its strong generative capability under sparse input conditions. MAGE achieves the highest accuracy while simultaneously reducing both MPJVE and Jitter. This indicates that the generated results are not only more natural, as evidenced by the reduced Jitter, but also capture more precise dynamic information, as reflected by the lower MPJVE. These improvements comprehensively enhance the model’s ability to capture both spatial and temporal information, from the process to the final results.\par
We also observe a negative correlation between static accuracy and dynamic coherence. When the focus is overly localized, it is easier to improve the accuracy of individual frames at the expense of dynamic continuity across the sequence. Conversely, focusing on the overall sequence can improve continuity while sacrificing local accuracy. MAGE addresses this inherent trade-off by introducing a multi-stage denoising strategy that balances local accuracy and sequence coherence, which is especially valuable.\par
Dataset $\text{D}_2$ uses a larger training set and employs a different dataset for testing, resulting in distinct data distributions in the training and test sets. This setup places greater emphasis on the model’s capacity for generalization and transfer. The results in Table~\ref{tab:protocol2} indicate that MAGE performs strongly on $\text{D}_2$, achieving the best outcomes on the MPJRE, MPJVE, and Jitter metrics. It also surpasses the two other diffusion-based algorithms in MPJPE and ranks second only to AvatarJLM~\cite{zhengRealisticFullBodyTracking2023} among methods using three-point sparse inputs. Particularly remarkable is MAGE’s performance on the Jitter metric. Given that real data has a Jitter of 2.92, MAGE delivers a substantially lower Jitter than other baseline methods, surpassing the second-best performer, SAGE~\cite{fengStratifiedAvatarGeneration2024}, by 31.4\%.\par
\begin{table*}[t]
    \centering
    \begin{tabular}{lrrrrrrrr}
        \toprule
        Method  & MPJRE & MPJPE & MPJVE&HandPE&UpperPE&LowerPE&RootPE&Jitter \\
        \midrule
        Final IK \cite{rootmotionFinalIK2018}    & 16.77    & 18.09    &59.24&-&-&-&-&- \\
        LoBSTr \cite{yangLoBSTrRealtimeLowerbody2021a}       & 10.69    & 9.02     &44.97&-&-&-&-&-    \\
        VAE-HMD \cite{dittadiFullBodyMotionSingle2021}      &4.11      &6.83      &37.99&-&-&-&-&-\\
        Avatorposer \cite{jiangAvatarPoserArticulatedFullBody2022a}&3.08&4.18&27.70&2.12&1.81&7.59&3.34&14.49\\
        AvatarJLM \cite{zhengRealisticFullBodyTracking2023}&2.90&3.35&20.79&1.24&1.42&6.14&\underline{2.94}&8.39\\
        AGRoL \cite{duAvatarsGrowLegs2023}&2.66&3.71&\underline{18.59}&1.31&1.55&6.84&3.36&7.26\\
        SAGE \cite{fengStratifiedAvatarGeneration2024}&\underline{2.53}&\underline{3.28}&20.62&\underline{1.18}&\underline{1.39}&\underline{6.01}&2.95&\underline{6.55}\\
        Ours&\textbf{2.40}&\textbf{3.21}&\textbf{16.71}&\textbf{1.02}&\textbf{1.32}&\textbf{5.93}&\textbf{2.89}&\textbf{6.27}\\
        \bottomrule
    \end{tabular}
    \caption{Comparison of our method with some state-of-the-art methods on $\text{D}_1$. MAGE outperforms other methods and achieves the best performance on MPJPE [cm], MPJRE [deg], MPJVE [cm/s], Jitter [\text{10\textsuperscript{2}m/s\textsuperscript{3}}] metrics. In PE of local regions, MAGE also have state-of-the-art performance. The results shows that MAGE increases both the accuracy and continuity of the generative results. }
    \label{tab:pro1}
\end{table*}
\begin{figure*}[t]
    \centering
    \includegraphics[scale=0.52]{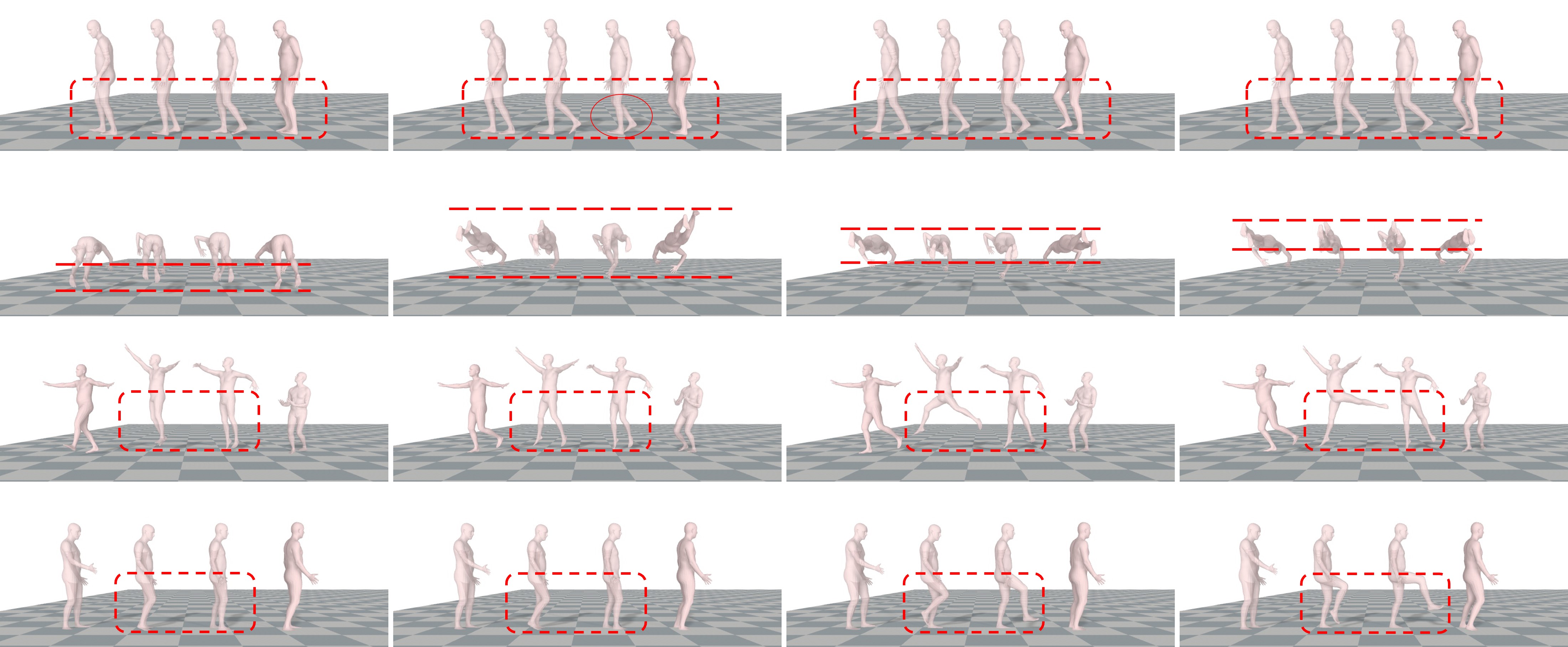}    
    \begin{tabular}{p{0.225\textwidth}p{0.225\textwidth}p{0.225\textwidth}p{0.225\textwidth}}
        \centering AGRoL  & 
        \centering SAGE  & 
        \centering Ours  & 
        \centering GT 
    \end{tabular}
    \caption{Visualization results of typical motions compared with other methods under $\text{D}_1$. From top to bottom: backward walking, freestyle swimming, ballet dancing, and kicking.}
    \label{visualization}
\end{figure*}

Figure~\ref{visualization} presents several visualization results of MAGE and other baseline methods under $\text{D}_1$. We selected four representative movements—backward walking, freestyle swimming, ballet dancing, and kicking—to visualize the performance of each method. Overall, our results show that MAGE clearly outperforms AGRoL and SAGE. Specifically, for backward walking, AGRoL tends to underestimate the stride, making the movements appear smaller than they are, while SAGE exhibits noticeable errors in the positions of the left and right feet. In contrast, MAGE’s predictions closely match the ground truth. Freestyle swimming poses a particular challenge in predicting leg motion, because the flutter kick of the lower legs can be largely independent of the upper body’s paddling. Therefore, we focused our evaluation on the approximate leg positions and the range of foot movements. Here again, MAGE delivers more stable predictions and shows greater accuracy in capturing global position, highlighting the benefits of the multi-scale design. Finally, in ballet dancing and kicking, where leg movements can be very large in range, AGRoL and SAGE struggle to reconstruct the lower body accurately. In comparison, MAGE performs significantly better and basically reconstructs the correct movements, further demonstrating its effectiveness.
\begin{table}[t]
    \centering
    \begin{tabular}{lrrrrrrrr}
        \toprule
        Method  & MPJRE & MPJPE & MPJVE&Jitter \\
        \midrule
        VAEHMD\textsuperscript{†}     & -    & 7.45    &-&- \\
        FLAG\textsuperscript{†}      & -    & \underline{4.96}    &-&- \\
        AvatarPoser& 4.70    & 6.38    &34.05&10.21 \\
        AvatarJLM& \underline{4.30}    & \textbf{4.93}    &26.17&7.19 \\
        AGRoL& \underline{4.30}    & 6.17    &\underline{24.40}&8.32 \\
        SAGE&4.62&5.86&33.54&\underline{7.13}\\
        Ours&\textbf{4.26}&5.60&\textbf{22.59}&\textbf{5.81}\\
        \bottomrule
    \end{tabular}
    \caption{Comparison of our method with some state-of-the-art methods on $\text{D}_2$. Methods with † use position and rotation of pelvis as an additional input, which are not directly comparable. The result of AvatarPoser is provided by \protect\cite{duAvatarsGrowLegs2023}.}
    \label{tab:protocol2}
\end{table}
\subsection{Ablation Study}

\begin{table}[t]
    \centering
    \begin{tabular}{lrrrr}
        \toprule
        Scale Set & MPJRE & MPJPE & MPJVE&Jitter \\
        
        \midrule
        $\mathbf{S}_3$  & 2.51   &3.39  &18.81&8.34 \\
        $\mathbf{S}_1,\mathbf{S}_3$ & \underline{2.43} & 3.26 & 18.66 & 9.62 \\
        $\mathbf{S}_2,\mathbf{S}_3$ & 2.44 & 3.30 & 18.72 & 9.79 \\
        $\mathbf{S}_1,\mathbf{S}_2,\mathbf{S}_3$ & \textbf{2.40} & \underline{3.21} & \textbf{16.71} & \textbf{6.27} \\
        $\mathbf{S}_0,\mathbf{S}_1,\mathbf{S}_2,\mathbf{S}_3$ & \textbf{2.40} & \textbf{3.18} & \underline{16.79} & \underline{6.80} \\
        \bottomrule
    \end{tabular}
    \caption{Ablation study on the use of scale set in MAGE. $\mathbf{S}_0$ denotes a single node representing the entire body motion.}
    \label{scaleset}
\end{table}
\begin{table}[t]
    \centering
    \begin{tabular}{lrrrrrrrr}
        \toprule
        Architecture  & MPJRE & MPJPE & MPJVE&Jitter \\
        \midrule
        Sequential  & 2.94    & 4.18&31.93&17.33 \\ 
        Gradual     & 2.40    & 3.21    &16.71& 6.27 \\
        \bottomrule
    \end{tabular}
    \caption{Ablation study of our diffusion-based model's architecture on $\text{D}_1$.  }
    \label{structure_ablation}
\end{table}
\begin{table}[t]
    \centering
    \begin{tabular}{lrrrrrrrr}
        \toprule
        Fusion Method  & MPJRE & MPJPE & MPJVE&Jitter \\
        \midrule
        $\textbf{C}+\textbf{F}$    &\textbf{2.40}&  \underline{3.22}   & \underline{17.92}    &\underline{8.86} \\
        $\textbf{C}+\textbf{F}_{rec}$  & \underline{2.44}    & 3.24    &18.41&9.42 \\
        $\textbf{C}+\textbf{F}+\textbf{F}_{rec}$    &\textbf{2.40} & \textbf{3.21}    & \textbf{16.71}    &\textbf{6.27} \\
        \bottomrule
    \end{tabular}
    \caption{Ablation of the fusion method for intermediate output. We concatenate the features and employ a fully connected layer to map them to the latent dimension. 
    }
    \label{fusion ablation}
\end{table}
In this section, we conduct ablation experiments under $\text{D}_1$ to demonstrate the effectiveness of our method.\\
\textbf{Scale Set.} We conduct ablation experiments with different scale combinations. As shown in Table~\ref{scaleset}, when more scale levels are used during the generative process, MAGE is better able to capture spatial and temporal information, leading to improved generation results. However, the performance of the four-stage training with the inclusion of $\mathbf{S}_0$ actually worsens. This is because the process of reconstructing a single node from three input nodes can not provide motion information based on the human body structure. While it offers a stronger global position constraint, the errors introduced at this stage negatively impact the rotation and motion details of body parts.\\
\textbf{Architecture.} We evaluate two kinds of diffusion-based architectures that can implement multi-stage generation. In sequential architecture, the model consists of three tandem diffusion parts, each responsible for a specific phase of the generative task. In contrast, gradual architecture divides a single diffusion model into three phases and cascades the generation results to realize its function. As shown in Table~\ref{structure_ablation}, the sequential diffusion architecture performs worse due to the introduction of additional noise, which enhances the diversity of results. However, as a generative task aimed at reconstructing the ground truth, the increased diversity is unnecessary and makes a detrimental effect. By contrast, the gradual method using a single diffusion model achieves state-of-the-art results. For a fair comparison, two models use the same denoising module with the same total number of layers.\\
\textbf{Fusion Method.} We also investigate how MAGE fuses features at each stage to guide later training. Specifically, we explore the latent sparse observations $\mathbf{C}$, the intermediate stage output $\mathbf{F}$, and the recovered latent human motion features $\mathbf{F}_{rec}$. According to Table~\ref{fusion ablation}, using only $\mathbf{C}$ and $\mathbf{F}_{rec}$ unavoidably loses some crucial information from $\mathbf{X}_{t}$, leading to acceptable results on the training set but lowest performance on the test set. In contrast, combining all three features $\mathbf{C}$, $\mathbf{F}$ and $\mathbf{F}_{rec}$ introduces additional constraints that improve generation quality.\\

\section{Conclusion}
In this paper, we investigate the problem of generating 3D human motion sequences based on sparse obsevations. To this end, we introduced a multi-scale human motion representation and proposed a multi-stage conditional diffusion model, MAGE, which progressively generates motion sequences in a coarse-to-fine manner. At each scale, the partially generated motion sequence not only supervises the training process but also acts as a new condition for guiding subsequent denoising and refinement. Our extensive experiments on publicly available datasets demonstrate that MAGE achieves state-of-the-art results, effectively balancing accuracy and continuity. Moreover, by decomposing the generation process across multiple scales, our approach provides a flexible framework for integrating additional constraints or priors in future extensions.

\newpage
\bibliographystyle{named}
\bibliography{reference}

\end{document}